\documentclass{article}

\usepackage{PRIMEarxiv}

\usepackage{amsmath,amssymb,amsfonts}  
\usepackage[utf8]{inputenc} 
\usepackage[T1]{fontenc}    
\usepackage{hyperref}       
\usepackage{url}            
\usepackage{booktabs}       
\usepackage{amsfonts}       
\usepackage{nicefrac}       
\usepackage{microtype}      
\usepackage{lipsum}
\usepackage{fancyhdr}       
\usepackage{graphicx}       
\graphicspath{{media/}}     

\usepackage{multirow}        
\usepackage{caption}
\usepackage{subcaption} 
\usepackage{listings} 
\usepackage{xcolor}   

\usepackage{pifont} 

\newcommand{\cmark}{\ding{51}}  
\newcommand{\xmark}{\ding{55}}  

\title{DINO-MX: A Modular \& Flexible Framework for Self-Supervised Learning
}

\author{
  Mahmut S. Gokmen \\
  Computer Science Department \\
  University of Kentucky \\
  Lexington, KY, USA\\
  \texttt{m.gokmen@uky.edu} \\
  \And
  Cody Bumgardner \\
  Computer Science Department \\
  University of Kentucky \\
  Lexington, KY, USA\\
  \texttt{cody@uky.edu} \\
}
\begin{document}
\maketitle

\begin{abstract}
Vision Foundation Models (VFMs) have advanced representation learning through self-supervised methods. However, existing training pipelines are often inflexible, domain-specific, or computationally expensive, which limits their usability across different domains and resource settings. DINO-MX is a modular and extensible training framework that combines the core principles of DINO, DINOv2 and DINOv3 within a unified configuration-driven system. It supports a variety of transformer-based architectures and is fully compatible with the Hugging Face ecosystem. The framework includes multiple training strategies such as low-rank adaptation (LoRA), layer freezing, and knowledge distillation, along with support for distributed training through both Distributed Data Parallel (DDP) and Fully Sharded Data Parallel (FSDP). DINO-MX is designed to work with both natural and specialized data types, including single- and multi-channel images. Experimental results on diverse datasets show that DINO-MX achieves competitive performance while significantly reducing computational costs. Additionally, it offers interpretability tools and a label-guided data augmentation method that improves attention-based localization without the need for extra detection or segmentation heads. DINO-MX provides a reproducible and scalable foundation for developing, adapting, and benchmarking self-supervised vision models across a range of research and real-world applications.
\end{abstract}

\section{Introduction}

The significant improvements in deep learning architectures in recent years have made foundational models widely available for everyday use through Large Language Models (LLMs), Vision Language Foundational Models (VLFMs), Vision Foundational Models (VFMs), and Multi-modal Foundational Models (MMFMs). These models leverage the large-scale datasets they've been trained on to serve as generators of generalized results across diverse domains. The remarkable capacity of these models to understand complex patterns and semantic relationships has transformed numerous fields, from natural language processing to computer vision \cite{LLM_review, survey_vlfm_outlook, vfm_review1}.

In recent years, released VFM training techniques relying on Self-Supervised approaches have led to substantial improvements in medical imaging analyses. These approaches have demonstrated particular promise for tasks ranging from anatomical structure segmentation to disease classification and anomaly detection \cite{vfm_class, vfm_seg}. Despite these advances, a widely adopted framework for the development, training, and optimization of medical vision foundation models remains absent. Vision foundation models are characterized as self-supervised models trained on large-scale datasets that learn generalizable visual representations transferable to multiple downstream tasks \cite{vfm_class_2}. Self-supervised learning (SSL) represents a training paradigm where models learn representations without explicit labels by solving pretext tasks derived from the data itself.

Vision foundation models like DINO and DINOv2 \cite{DINO_paper, DINOv2} undoubtedly implement significant advancements in general computer vision applications. Their self-supervised training methodologies have proven remarkably effective at learning rich, semantically meaningful visual representations from diverse datasets. It seems reasonable to assume that for medical imaging tasks, these models' feature representation capacity would transfer effectively. However, this is not always the case. Consider a foundation model architecture where the model has been pre-trained on natural images but must be adapted to specialized medical domains such as CT scans or histopathology slides \cite{prov-gigapath, domain_adapt_med}. In this situation, the domain gap emerges as a function of fundamental differences in data distributions, including pixel intensity ranges, spatial relationships, and semantic content. For instance, the pixel value distributions in natural RGB images have fundamentally different statistical properties than Hounsfield units in CT scans, necessitating specialized adaptation techniques. This discrepancy often results in suboptimal performance when models trained on natural images are directly applied to medical tasks without appropriate adaptation strategies \cite{Are_Natural_Domain}.

Despite the existence of vision foundational models closing the gap between medical image domain and natural image domain, it is still necessary to adapt these pretrained models in desired research fields by using cost-efficient approaches like parameter efficient fine-tuning, low rank adaptation or standard fine-tuning approaches. The majority of released vision foundational models especially trained with large-scale medical datasets still use non-standardized backbone architectures or Vision Transformers (ViT). Researchers working in this field must tackle this issue if they want to distill or fine-tune their pre-trained models. Especially, clinical institutes with limited resources face challenges implementing these pre-trained models without sufficient knowledge about these architectures, since a standardized framework is not released for the most well-known foundational models. This represents a significant obstacle to the generalizability of vision foundational models due to the lack of a generalized framework that every institute can use easily. The implementation complexities often require specialized expertise in deep learning architectures, optimization techniques, and computational infrastructure, creating barriers to entry for many potential users in the medical community. Missing generalizability also causes poor performance on real clinical data when considering that all these released pre-trained models are trained on publicly available datasets, which may not fully represent the variability and complexity encountered in real-world clinical environments.

In the best scenario, such as having enough resources to pre-train a foundational model using a large-scale dataset, it should be considered that the training techniques developed by scientists are adapted for natural images. These techniques often incorporate assumptions about image statistics, feature distributions, and semantic hierarchies that may not hold true for medical imaging. Depending on which special type of medical dataset the model will be trained on, data augmentation techniques play a crucial role in training progress \cite{data_aug_1, data_aug_2}. The strategic selection and implementation of appropriate augmentation methods can significantly impact model performance, generalization capabilities, and robustness to variations in imaging conditions. As it stands, the data augmentation techniques released within training frameworks are also developed for natural images. Although numerous data augmentation techniques have been developed for medical data processing for deep learning and machine learning approaches, current frameworks specifically developed for generating representations of medical images still provide data augmentation techniques optimized for natural images. This misalignment between augmentation strategies and domain-specific requirements can limit the effectiveness of the resulting models. Furthermore, one of the most popular and widely used data types (CT scans with single-channel input) are not compatible with foundational models optimized for CT scan image representation generation unless CT slices are artificially multiplied to three channels \cite{CT_CLIP}. This artificial channel multiplication represents a workaround rather than a principled solution, potentially introducing artifacts or distortions in the resulting representations.

Another issue that should be solved is efficient training. Even with foundational models utilized in pre-training, implementation of standard foundational model training can potentially take more than 2 days when trained on multiple nodes; in circumstances using a single node, that consumption time increases in direct proportion. The computational intensity of training large-scale vision foundation models poses significant challenges, particularly for researchers with limited access to high-performance computing resources. Thus, parameter efficient fine-tuning (PEFT) plays a crucial role in pre-training and fine-tuning tasks to generate representations capturing the most important features from medical images \cite{peft_paper, vfm_review1}. PEFT strategies can dramatically reduce the computational footprint while maintaining model performance, making foundation model adaptation more accessible and sustainable. Furthermore, considering the higher resolution of medical images compared to natural images, PEFT is essential for pre-training or fine-tuning a foundational model. The increased spatial dimensions in medical images like high-resolution CT or MRI scans multiply the computational demands exponentially, making efficient adaptation strategies not just desirable but necessary. Fortunately, recently introduced PEFT techniques are highly efficient when applied to desired models. The problem actually arises during application. Most recent frameworks released within the last two years lack PEFT or Low Rank Adaptation capabilities for fine-tuning \cite{DINO_paper,DINOv2,survey_vlfm_outlook}. This limitation creates significant implementation barriers for researchers without extensive programming expertise or computational resources. Researchers or practitioners working with these models must apply PEFT techniques separately or modify existing models while being cautious not to break them. These modifications may require low-level changes that can cause errors when loading pre-trained models, introducing potential instabilities or compatibility issues that further complicate the adaptation process.

Another approach, known as distillation training, requires less memory than PEFT techniques and helps distill different-sized models even when they don't share the same architecture \cite{distill_paper}. Knowledge distillation offers a promising path for transferring capabilities from large, computation-intensive models to smaller, more deployable variants that maintain much of the original performance. This technique helps researchers extend their studies by distilling larger models into smaller, task-specific foundational models that can be more easily deployed in resource-constrained clinical environments \cite{knowledge_dist}. Although distillation seems like a solution for training frameworks that don't include PEFT techniques, distillation training requires a different approach than standard foundational model training techniques, necessitating a different working framework. The implementation of effective distillation strategies requires careful consideration of teacher-student architectures, loss functions, and training protocols that differ substantially from standard fine-tuning approaches. Notably, training frameworks such as DINO and DINOv2 don't offer PEFT or distillation techniques in their implementations, despite being among the most used training techniques for foundational models in recent years \cite{DINOv2, DINO_paper}. This limitation constrains the practical utility of these otherwise powerful frameworks for medical imaging applications, particularly in settings where computational efficiency is paramount.

All the problems mentioned above represent significant challenges we have faced while working with foundational models in medical imaging contexts. Undoubtly, the frameworks exist and released as an open source considering these challenges. Even these frameworks are actively using by researchers, people will have to deal with the problems and limitations as they have listed above. As a part of this paper, the DINO-MX framework will be presented. It is important to note that our purpose in developing this framework is not to achieve optimal results for any specific model on a particular dataset or to focus on a singular task. Rather, our aim is to provide a comprehensive framework that researchers in this field can benefit from, especially medical laboratories, enabling them to save time and reach accurate results efficiently. In today's landscape, state-of-the-art AI models already demonstrate exceptional performance, and achieving near-optimal results is virtually inevitable when utilizing high-capacity GPUs. The true innovation in our proposed solution lies in providing an infrastructure where researchers can quickly compare the latest open-source models and evaluate results without unnecessary time expenditure, thus accelerating the pace of medical imaging research and implementation.

To further motivate our approach, we will now detail the critical limitations that hinder the widespread adoption of foundational models in medical imaging.

\subsection{Domain Gap Between Natural and Medical Images}
The foundational models provided by major commercial companies such as Meta, Google, and OpenAI are primarily trained on natural image datasets like ImageNet or LAION. While numerous medical foundation models have recently been introduced by academic labs and research consortia, a significant portion of them still rely on pretrained weights obtained from vision models originally trained on natural domain data. This reliance is due to the computational and resource constraints of training large-scale models from scratch using medical data \cite{mia_challenge_1, mia_challenge_2}.

\begin{figure}[htbp]
    \centering
    \begin{subfigure}{0.48\textwidth}
        \centering
        \includegraphics[width=\textwidth]{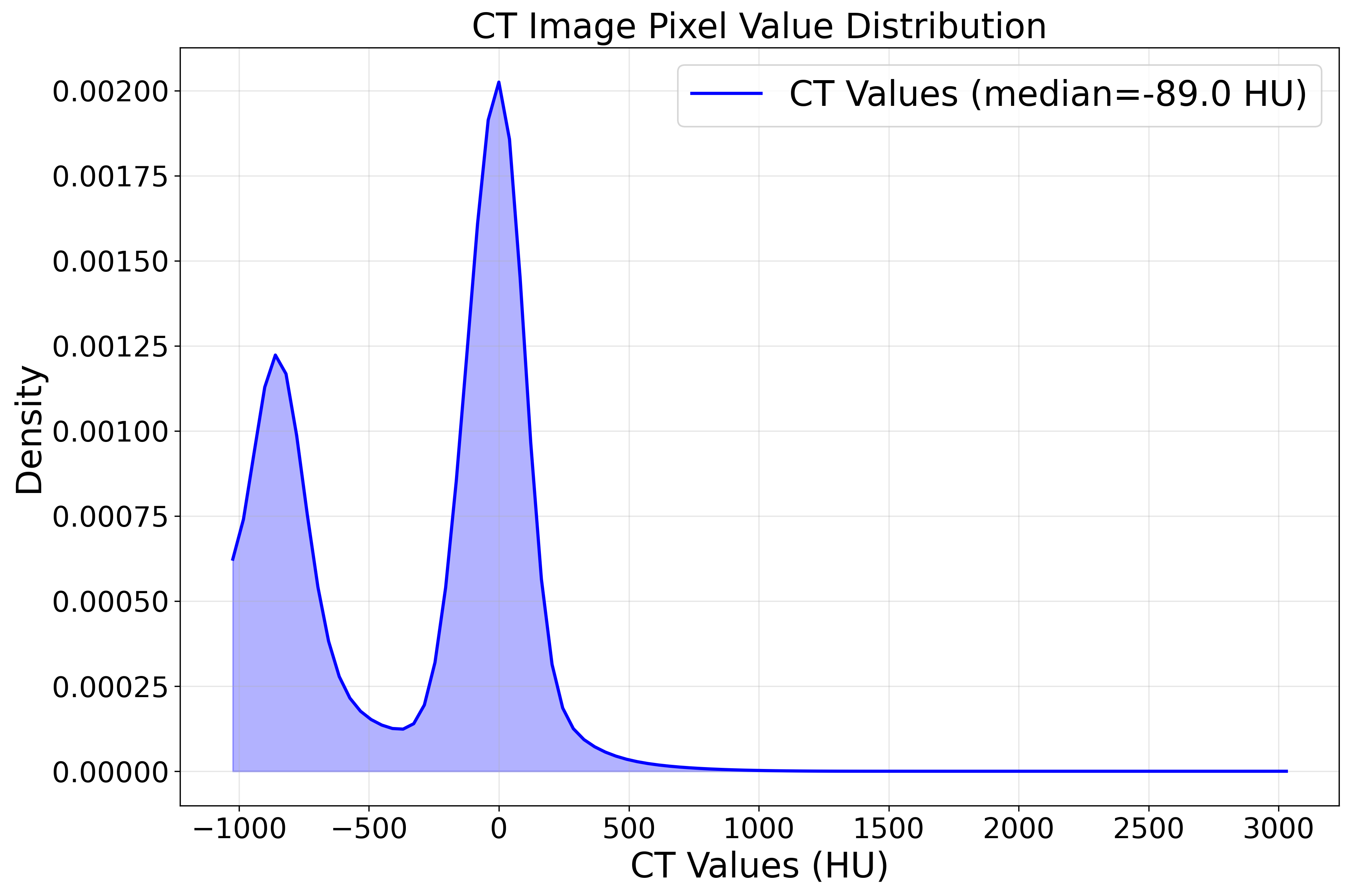}
        \caption{Pixel value distribution of a CT scan image (in Hounsfield Units). The histogram shows a typical peak around 0 HU, representing soft tissue, and a smaller peak in the air range (around -1000 HU).}
        \label{fig:fig1_sub1} 
    \end{subfigure}
    \hfill
    \begin{subfigure}{0.48\textwidth}
        \centering
        \includegraphics[width=\textwidth]{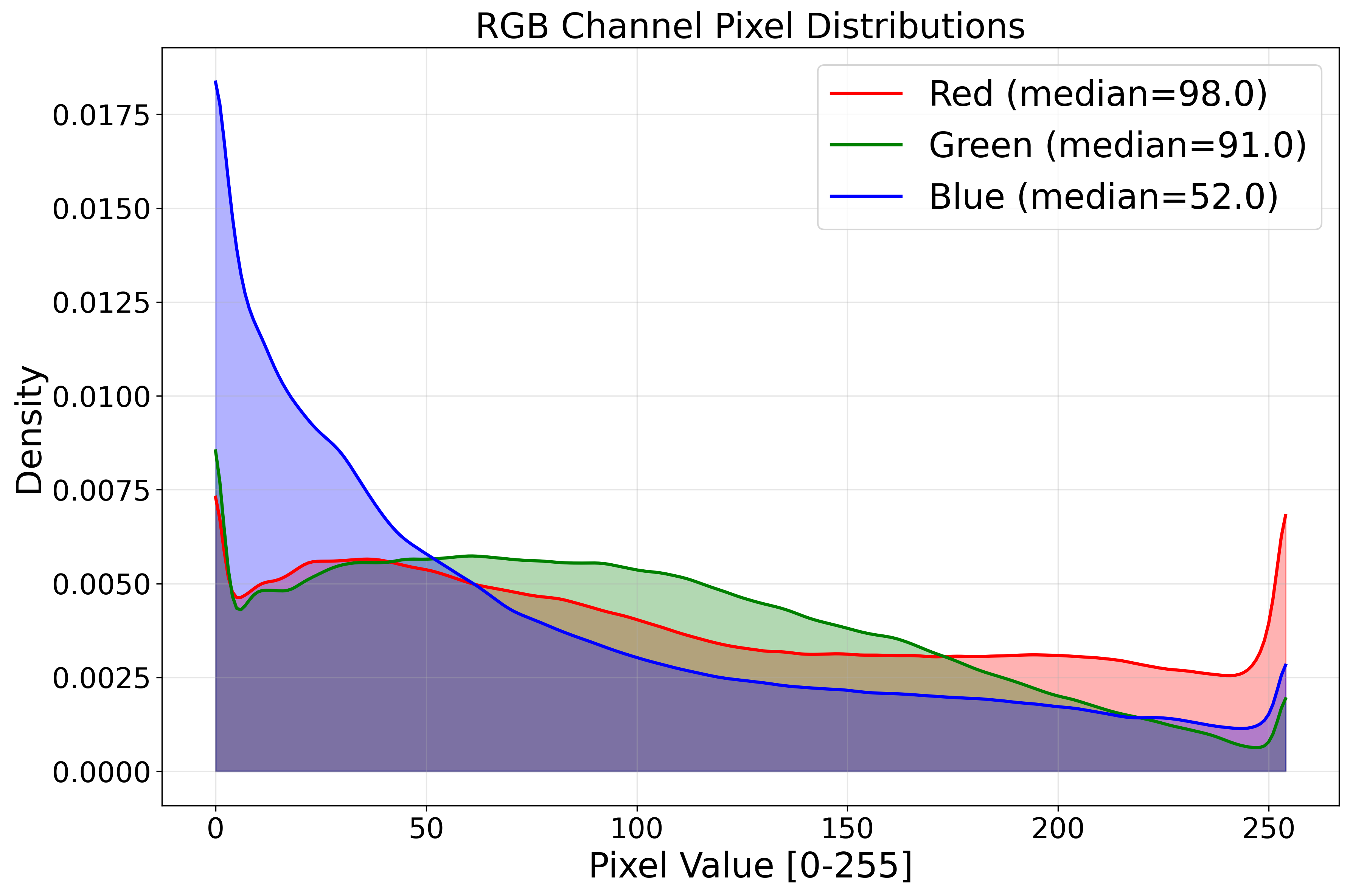}
        \caption{RGB channel-wise pixel value distributions for a natural image dataset. Each color channel displays a different pixel intensity distribution, commonly centered near higher values.}
        \label{fig:fig1_sub2} 
    \end{subfigure}
    \caption{Comparison of pixel value distributions between (a) a CT scan image and (b) a natural image. The CT distribution is defined in Hounsfield Units (HU), while natural images are represented in RGB channels with pixel values in the [0-255] range.}
    \label{fig:two-figures}
\end{figure}

However, natural images and medical images differ substantially in terms of data distribution, visual semantics, and imaging modalities. Medical images are often acquired through specialized techniques such as MRI, CT, PET, and histopathology, each governed by distinct physical principles and clinical protocols. These images vary across anatomical regions, resolutions, and even across institutions due to scanner heterogeneity. As highlighted in the literature \cite{domain_adapt_med} and illustrated in Figure \ref{fig:two-figures}, this domain gap poses a critical limitation for directly applying natural image models to medical tasks without adaptation.

Recent studies have shown that models pretrained on natural images can still achieve state-of-the-art performance on medical tasks when combined with appropriate training strategies and fine-tuning approaches. Nevertheless, the question remains: would foundation models specifically designed and pre-trained for medical data outperform their natural-image counterparts? This uncertainty continues to motivate the development of domain-specific models tailored to the unique challenges of medical imaging \cite{Are_Natural_Domain}.

\subsection{Generalizability Issues}
Despite promising advancements in vision foundation models for medical imaging, significant challenges remain regarding their generalizability across diverse clinical scenarios. These models often demonstrate exceptional performance on the specific datasets they were trained on but show marked degradation when exposed to data from different institutions, equipment manufacturers, or patient populations \cite{fed_found_models}. This phenomenon, commonly referred to as the "dataset shift" problem, arises from variations in image acquisition protocols, scanner parameters, and institutional imaging practices that introduce distributional differences between training and deployment environments \cite{x_ai}.

The ICOVAI study \cite{ai_generalizability} provides compelling empirical evidence of these generalizability challenges, as shown in Table \ref{tab:icovai_performance}. Despite being developed on a large multicenter dataset (n = 1,286 CT scans), the model's performance decreased significantly during external validation. Infectious lung opacity segmentation performance dropped from a DSC of 0.76 to 0.59, while COVID-19 classification agreement fell from kappa = 0.78 to 0.62 (both p < 0.0001). Notably, the model maintained excellent performance on basic lung contour segmentation (DSC = 0.97) across both test sets, suggesting that generalizability issues are more pronounced for complex, disease-specific tasks most relevant to clinical decision-making.

\begin{table}[htbp]
\centering
\caption{Performance Comparison of ICOVAI AI Model on Internal vs. External Validation. (Not: 'tablenotes' için 'threeparttable' paketi gerekebilir.)}
\label{tab:icovai_performance}
\begin{tabular}{lccc}
\toprule
\textbf{Performance Metric} & \textbf{Internal Test} & \textbf{External Test} & \textbf{Significance} \\
\hline
\multicolumn{4}{l}{\textit{Segmentation Performance (Dice Similarity Coefficient)}} \\
Lung Contours & 0.97 & 0.97 & n.s. \\
Infectious Lung Opacities & 0.76 & 0.59 & p < 0.0001 \\
\hline
\multicolumn{4}{l}{\textit{Classification Performance}} \\
CO-RADS Agreement (Cohen's kappa) & 0.78 & 0.62 & p < 0.0001 \\
\hline
\multicolumn{4}{l}{\textit{Dataset Characteristics}} \\
Sample Size (CT scans) & 1,286 & 400 & - \\
\bottomrule
\end{tabular}
\parbox{\linewidth}{\small 
CO-RADS = COVID-19 Reporting and Data System; n.s. = not significant. This table demonstrates the significant performance drop in both infectious lung opacity segmentation and COVID-19 classification tasks when the model was externally validated...
}
\end{table}

\subsection{High Computational Cost}
Building foundation models requires enormous computing power and money. Recent studies show that training costs for large AI models have grown by 2.4 times each year since 2016, with major models like GPT-4 and Gemini Ultra costing between \$78-191 million for a single training run \cite{ai_cost_1}. If this trend continues, by 2027, training the largest models will cost over \$1 billion, limiting this capability to only the richest organizations. These costs include specialized computer hardware (about 50\%), skilled researchers and engineers (30-50\%), and additional expenses for supporting equipment and electricity \cite{ai_cost_1}.

When adapting these general models for medical imaging, costs rise even higher. Medical data requires special expertise and more complex model designs to handle different types of scans like X-rays, CT scans, and MRIs. Although healthcare AI shows good financial returns-about \$3.20 for every \$1 invested-the high initial costs create major barriers to implementation \cite{ai_cost_2}. This especially affects research centers and hospitals with limited budgets, potentially widening the gap between wealthy and less wealthy healthcare systems worldwide. While techniques like transfer learning (adapting existing models instead of building from scratch) can help reduce costs, they still require significant computing resources and often result in performance compromises that may be unacceptable for critical medical applications.

\subsection{Insufficient Data Augmentation Strategies in Medical Foundational Models}
Data augmentation techniques play a crucial role in training vision foundational models, especially in scenarios where data is inadequate to train the model from scratch. Various augmentation techniques, as listed in Table \ref{tab:data_aug_table}, demonstrate that task-specific models routinely leverage predefined augmentation techniques designed for specific domains. These techniques prove invaluable when training task-specific deep learning models for segmentation, classification, and detection tasks \cite{data_aug_1, data_aug_2}.

\begin{table}[htbp]
\centering
\scriptsize
\caption{Representative Data Augmentation Techniques in Medical Imaging Tasks}
\label{tab:data_aug_table}
\begin{tabular}{p{2.2cm}|p{2cm}|p{3cm}|p{2cm}|p{3cm}}
\toprule
\textbf{Task} & \textbf{Modality} & \textbf{Augmentation Methods} & \textbf{Dataset} & \textbf{Performance Metrics} \\
\midrule
Hepatic vessel segmentation & CT & Resampling, feature extraction & MSD & DSC: 79\%, Sens: 82.2\%, Spec: 95.1\% \\
Hepatic lesion segmentation & MRI & Histogram equalization, z-score norm. & 48 MRI cases & DSC: 0.98, HD: 1.02 mm \\
Brain tumor classification & MRI & Resampling, normalization, geometric aug. & TCGA-LGG & F1: 92\%, Acc: 92\% \\
Alzheimer's classification & MRI & Format conversion, basic aug. & ADNI & Acc: 98.68\%, F1: 98.68\% \\
Chest nodule detection & X-ray & Flipping, rotation, scaling & Chest X-ray & MAE: 0.0296, PSNR: 28.98, SSIM: 0.7555 \\
Lung/colon cancer detection & Histopathology & Noise removal, patch extraction & LC25000 & Acc: 99.33\%, F1: 98.27\% \\
Head/neck tumor segmentation & CT & Transformations, intensity shift & TCIA HNSCC & DSC: 0.90 \\
Breast cancer classification & Histopathology & Augmentation, normalization & BreakHis, IDC, UCSB & Acc: 85.7\% \\
Breast lesion segmentation & Mammography & Translation, rotation, scaling & 204 CESM images & Dice: +10.2\% \\
Cardiac disease classification & Echo & Frame sampling, color conv. & 381 patients & AUC: 0.97, F1: >90\% \\
\bottomrule
\end{tabular}
\end{table}

From the perspective of foundational models, however, there is unfortunately no standardized data augmentation strategy specifically designed for datasets employed in the training phase. Most researchers and developers working on vision foundational models merely utilize standard data augmentation techniques that are pre-defined and released within their frameworks. This represents a significant gap for foundational models, particularly when considering that pre-trained models such as DINOv2 rely heavily on weights fully trained with natural images rather than domain-specific medical imagery.
Furthermore, the majority of studies involving pre-trained medical foundational models lack significant evidence regarding performance with different augmentation techniques, despite data augmentation being one of the key concepts in foundational model training. This limitation constrains the models' generalization capabilities and their adaptability to various medical imaging modalities.

\begin{figure}[htbp]
    \centering
    \includegraphics[width=\textwidth]{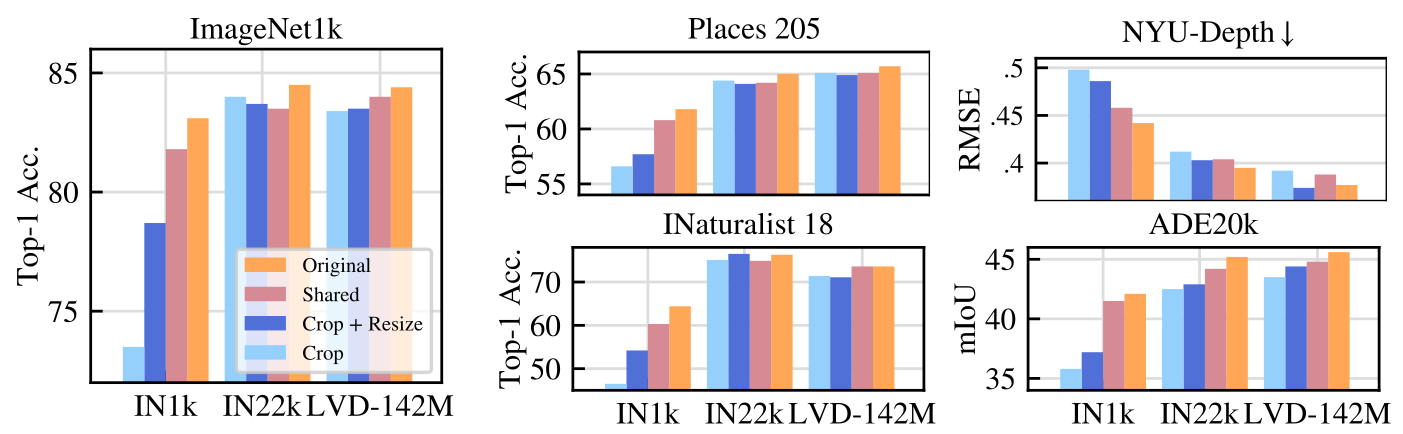}
    \caption{Impact of dataset size when varying data augmentations. Results of ViT-L on linear evaluation benchmarks. Cropping without resizing ('Crop') reaches very high performances comparable to full augmentation ('Original') on a wide variety of benchmarks when the dataset size is large enough.}
    \label{fig:augmentation_scaling}
\end{figure}

Although data augmentation techniques are crucial for fine-tuning and domain adaptation, the latest studies reveals that the necessity of data augmentation techniques has no significant effect on large scale datasets utilized on training with large models \cite{no_need_data}. The paper released by META, reveals that strong image representations can be obtained with joint-embedding architectures (JEAs) using only simple cropping without resizing when training data is sufficiently large. Their experiments with DINOv2 demonstrate that as dataset size increases from smaller collections like ImageNet-1k to larger ones like LVD-142M (142 million images), the performance gap between models trained with extensive augmentations and those trained with minimal augmentations nearly disappears. This challenges the prevailing belief that domain-specific augmentations are essential for self-supervised learning and suggests that scaling data, model size, and training duration can compensate for the absence of handcrafted augmentations.
Figure \ref{fig:augmentation_scaling} demonstrates this consistent pattern across five benchmark tasks.

Although recent studies suggest that large-scale datasets may reduce the need for complex data augmentation, the lack of standardized augmentation strategies for medical vision foundation models remains a key limitation. Most existing techniques are developed for natural images and may not transfer well to medical data due to differences in modality and visual structure.

\subsection{Absence of Unified Frameworks for Medical Foundation Models}
Frameworks are environments designed to support a wide range of tasks and facilitate model deployment in production settings. In image processing, various specialized libraries help with reading medical images \cite{pedregosa2011scikit}.

Beyond these specialized tools, building a foundation model typically requires core libraries such as PyTorch or TensorFlow. Recent statistics show that over 90\% of studies employ the PyTorch ecosystem to build AI models \cite{pytorch_tensorflow}. However, libraries for data handling and model construction are not sufficient on their own. Additional components such as loss functions, training loops, evaluation metrics, and optimization utilities are essential. Considering all these requirements, comprehensive open-source frameworks that provide end-to-end support are still limited.

One of the most widely used comprehensive frameworks is HuggingFace, which offers built-in models, training utilities, and a standardized model-sharing ecosystem. Many studies adapt their architectures to the HuggingFace format to increase visibility \cite{huggingface}. However, foundational model research often involves custom-built architectures that are not directly compatible, requiring additional engineering. Importantly, using the HuggingFace framework alone is not sufficient to train foundational models; researchers must still have a strong understanding of model training principles.

This raises a common question: why are many foundational models not developed entirely within frameworks like HuggingFace? The answer lies in the experimental flexibility required during early-stage development. Foundational model research often involves frequent low-level modifications, custom loss functions, and novel training strategies. As such, building models from the ground up using custom codebases is often more practical. Unfortunately, this approach has a downside: many vision AI studies remain non-reproducible or inaccessible due to a lack of public release \cite{reproduce}.

In response to these challenges, NVIDIA has introduced MONAI (Medical Open Network for AI), a high-level framework tailored for medical imaging \cite{monai}. While MONAI has gained momentum, its support for foundational models remains limited. Notably, popular modern training strategies such as DINOv1 and DINOv2 are not yet integrated into the MONAI ecosystem. Another recent framework, Lightly, incorporates many SSL training strategies (though it currently does not support DINOv2), but lacks integration with the medical imaging domain and provides no benchmark results on medical datasets, limiting its applicability in healthcare-related research \cite{lightly}.

\section{Methodology: The DINO-MX Framework}
\label{sec:methodology}

DINO-MX is a comprehensive and modular framework developed to enhance self-supervised learning (SSL), specifically tailored for Vision Transformer (ViT) architectures. It seamlessly integrates with widely-used model repositories such as Hugging Face, enabling standardized generation and deployment of model checkpoints. This standardized approach significantly reduces the implementation barriers typically encountered by researchers and practitioners, facilitating easier experimentation and broader adoption.

\begin{figure}[htbp]
\centering
\includegraphics[width=1\textwidth]{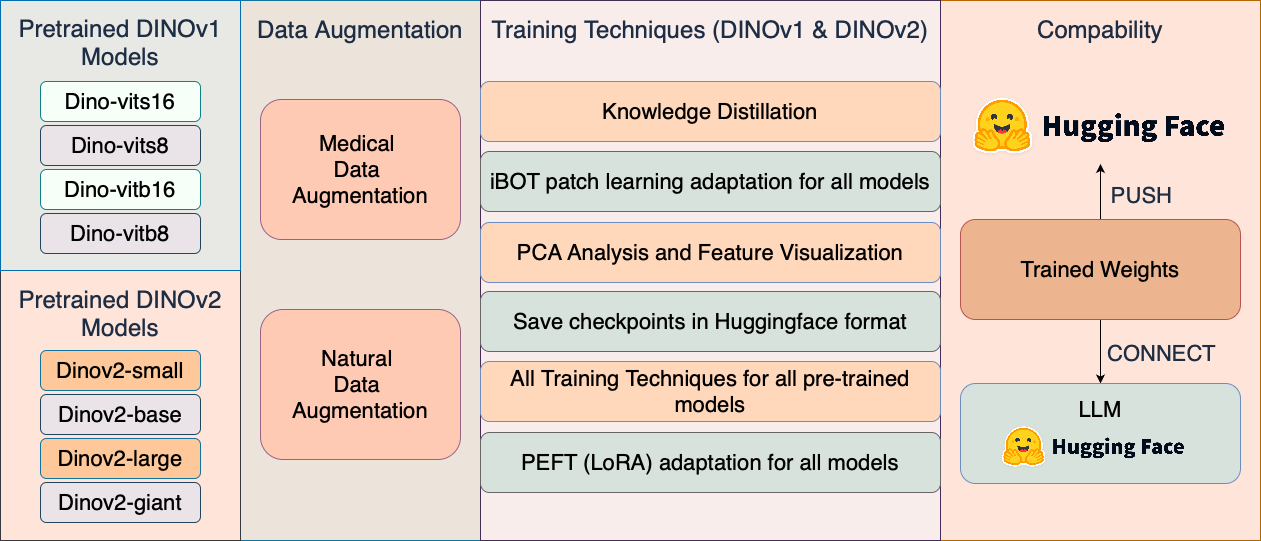}
\caption{General representation of DINO-MX framework}
\label{fig:dino_mx_general}
\end{figure}

Figure~\ref{fig:dino_mx_general} illustrates the overall structure and workflow of the DINO-MX framework, emphasizing its compatibility and interoperability features with Hugging Face. It showcases the modular approach from pre-trained models selection to data augmentation, training techniques, and final deployment.

Key features of the DINO-MX framework include:
\begin{itemize}
    \item \textbf{Standardized Backbone Integration:} Leveraging standardized ViT architectures ensures seamless compatibility with mainstream model repositories, simplifying model adoption and deployment.
    \item \textbf{Medical Image Adaptation:} Native support for single-channel medical images, such as CT scans, preserves original data integrity and avoids artificial multi-channel transformations.
    \item \textbf{Domain-Specific Augmentation:} Includes medical and natural image data augmentation strategies tailored for different modalities, effectively bridging domain gaps.
    \item \textbf{Parameter-Efficient Fine-Tuning (PEFT):} Incorporates efficient adaptation methods like Low Rank Adaptation (LoRA), significantly reducing the computational resources required for fine-tuning.
    \item \textbf{Model Distillation:} Facilitates knowledge transfer from large foundational models to smaller, more efficient models, supporting deployment in resource-limited clinical environments.
    \item \textbf{Modular Architecture:} Designed to enable researchers to easily mix, match, and extend components and training methodologies without extensive code modifications, fostering rapid experimentation and innovation.
    \item \textbf{Cross-Framework Compatibility:} Allows integration of different SSL approaches (e.g., DINOv1 and DINOv2), enabling novel cross-framework experiments and comparative analyses.
\end{itemize}

The subsequent sections provide detailed insights into specific training strategies implemented within the DINO-MX framework. These include parameter-efficient fine-tuning methods such as layer freezing, model distillation processes, LoRA-based adaptations, and attention map analyses for enhanced interpretability and performance evaluation. Further sections also discuss the modularity and flexibility of DINO-MX and validate its robustness and practical utility through extensive experimental evaluations across diverse medical imaging datasets.

\subsection{System Configuration and Execution}
DINO-MX framework uses two types of configuration files to run a training process. First configuration file designates accelerator attributes which identifies training strategy: FSDP and DDP.

\subsubsection{Parallelization Strategies}
Parallel computation is essential for training foundation models, as the scale of parameters and data volumes would make sequential processing prohibitively time-consuming. Modern vision transformers with billions of parameters demand distributed training approaches to achieve reasonable convergence times.

With advancements in GPU processing technology, parallel computation techniques have evolved and are primarily categorized into two approaches: Distributed Data Parallelism (DDP) and Fully Sharded Data Parallelism (FSDP).

\paragraph{Distributed Data Parallelism (DDP)} represents the conventional approach to multi-GPU training, where identical model copies are maintained across all devices, each processing different data batches. After computing gradients locally, an all-reduce operation synchronizes these gradients across devices before parameter updates. While DDP effectively scales training throughput, it requires each device to store a complete model copy, creating memory bottlenecks for extremely large architectures \cite{DDP_paper}.

\paragraph{Fully Sharded Data Parallelism (FSDP)}, a more recent innovation, addresses these memory limitations by sharding model parameters, gradients, and optimizer states across participating devices. During forward and backward passes, FSDP dynamically reshards parameters as needed, allowing much larger models to be trained on the same hardware \cite{FSD_paper}.

The DINO-MX framework implements both approaches using PyTorch's native distributed modules for DDP and FSDP integration.

In current frameworks such as DINOv1 and DINOv2, these parallel computation techniques are hardcoded, with DDP used for DINOv1 and FSDP for DINOv2. This limits researchers who want to evaluate model performance across different parallelization technologies. To overcome this limitation, the DINO-MX framework abstracts these complex configurations into a simple, high-level setting. This flexible approach allows researchers to run all types of DINO models with both parallelization techniques, enabling comprehensive performance benchmarking.

This abstraction significantly simplifies experimentation. As shown in Figure~\ref{fig:dist_config}, researchers no longer need to manage dozens of complex, backend-specific settings. Instead, they can switch between DDP and FSDP by changing a single \texttt{type} parameter. This modularity allows for the direct comparison of training efficiency and model performance across different parallelization approaches without any code modifications.

\begin{figure}[!ht]
\centering
\begin{lstlisting}[caption={
    The simplified distribution configuration in DINO-MX. Switching between
    FSDP and DDP is achieved by simply changing the 'type' parameter.
}, label=fig:dist_config]
distribution:
  type: fsdp # or ddp 
  mixed_precision: bf16
  downcast_bf16: 'no'
\end{lstlisting}
\caption{The simplified, high-level configuration for parallelization strategies in the DINO-MX framework.}
\label{fig:parallel_configs} 
\end{figure}

\subsubsection{Training Configurations}
The DINO-MX framework employs detailed training configuration files that control various aspects of the model training process. These configurations enable systematic experimentation with different training approaches while maintaining reproducibility.

The training configuration file is divided into several logical sections that control different aspects of the training process:

\begin{description}
    \item[DINO Head Configuration]
    Controls the representation learning parameters for the DINO self-supervised approach. The output dimension determines the size of the representation space, with larger dimensions potentially capturing more nuanced features.
    
    \begin{lstlisting}[caption=DINO head settings, label=lst:dino_head]
dino_head:
  out_dim: 65536  
  norm_last_layer: False
  loss_weight: 1.0
    \end{lstlisting}

    \item[iBOT Parameters]
    Configures masked image modeling for additional self-supervised pre-training objectives. The mask ratio parameters control the percentage of image patches randomly masked during training.

    \begin{lstlisting}[caption=iBOT parameters, label=lst:ibot]
ibot:
  loss_weight: 0.0
  out_dim: 65536 
  norm_last_layer: True
  mask_sample_probability: 0.5
  mask_ratio_min_max:
    - 0.1
    - 0.5
    \end{lstlisting}

    \item[Distillation Settings]
    Specifies parameters for transferring knowledge from larger to smaller models. It defines the teacher model, custom weights path, and loading method.
    
    \begin{lstlisting}[caption=Distillation settings, label=lst:distill]
distillation:
  distilled_model_type: 'facebook/dinov2-giant'
  distilled_model_weights: ''
  load_from_disk: False
    \end{lstlisting}

    \item[LoRA Configuration]
    Controls parameter-efficient fine-tuning using Low-Rank Adaptation. The rank parameter $r$ determines the dimensionality of the low-rank matrices.
    
    \begin{lstlisting}[caption=LoRA configuration, label=lst:lora]
lora_config:
  lora_r: 4
  lora_alpha: 16
  lora_dropout: 0.1
    \end{lstlisting}

    \item[Multi-crop Augmentation]
    Defines the data augmentation strategy using global and local views to develop scale-invariant representations.
    
    \begin{lstlisting}[caption=Multi-crop augmentation, label=lst:crops]
crops:
  global_crops_scale:
    - 0.4
    - 1.0
  local_crops_number: 8
  global_crops_number: 2
  local_crops_scale:
    - 0.1
    - 0.4
  global_crops_size: 224
  local_crops_size: 96
    \end{lstlisting}

    \item[Dataset Configuration]
    Specifies the dataset path and preprocessing parameters.
    
    \begin{lstlisting}[caption=Dataset configuration, label=lst:dataset]
dataset:
  dataset_path: dataset/pathmnist
  shuffle: true
    \end{sallallahu
}
    \end{lstlisting}

    \item[Core Training Parameters]
    Defines model selection, optimization settings, and general training hyperparameters. This section controls the teacher-student relationship, learning rates, PEFT flags, and checkpointing.
    
    \begin{lstlisting}[caption=Core training parameters, label=lst:train]
train:
  model_name: 'training_test_fsdp'
  model_type: 'facebook/dinov2-base'
  global_batch_size: 64
  max_iterations: 2000
  do_distillation: false
  mixed_precision: true
  use_lora: true 
  freeze_last_layer: 100
  warmup_teacher_temp_iterations: 500
  warmup_iterations: 1000
  teacher_temp: 0.04 
  freeze_backbone_layers: 0
  momentum_teacher: 0.996
  centering: 'centering'
  ibot_separate_head: false 
  use_pretrained: True
  generate_samples: True 
  lr: 1e-4
  min_lr: 1e-5 
  saveckp_freq: 250
    \end{lstlisting}
\end{description}

The modular design of the configuration file mirrors the overall modular architecture of the DINO-MX framework, facilitating experimentation with different self-supervised learning approaches and adaptation to various medical imaging tasks.

These configurations can be actively used together with the previously defined FSDP and DDP training strategies, providing researchers with comprehensive control over both computation distribution and learning objectives. The resulting combination of training options and parallelization strategies offers a wide spectrum of experimental possibilities for comparative performance studies across different model architectures, adaptation techniques, and medical imaging domains. This flexibility allows for systematic exploration of the efficiency-accuracy trade-offs that are critical when deploying vision foundation models in resource-constrained clinical environments.
\subsubsection{Execution Workflow}
After preparing the accelerator and training configuration files, a single command initiates the entire workflow:

\begin{verbatim}
accelerate launch --config_file configs/accelerator/fsdp_accelerator_config.yaml 
    dino_training/train_dino_rgb_fsdp.py 
    --train_config_file configs/dino/training.yaml
\end{verbatim}

This command combines the PyTorch Accelerate launcher, the accelerator configuration (FSDP or DDP), and the training configuration. During training, the framework provides comprehensive real-time monitoring:

\begin{verbatim}
2025-04-01 01:03:24,093 - dino_trainer_rank0 - INFO - 
Total Loss: 13.6886 Local DINO: 9.8890 Global DINO: 1.2194 
iBOT: 2.5802 LR: 0.000002 Weight Decay: 0.040057 
Teacher momentum: 0.996001 Current Batch Size: 64 
Iteration: 16/2000 Gpu ID: 0 Memory: 22.96/47.32
\end{verbatim}

These detailed logs provide critical insights, displaying:
\begin{itemize}
    \item \textbf{Component-wise Loss Breakdown}: Separate tracking of local DINO, global DINO, and iBOT losses.
    \item \textbf{Optimization Parameters}: Current learning rate, weight decay, and teacher momentum.
    \item \textbf{Resource Utilization}: GPU ID, memory consumption (used/total), and batch size.
    \item \textbf{Progress Indicators}: Current iteration count and timestamp.
\end{itemize}

Any changes to the training configuration are compatible with both FSDP and DDP. The framework supports switching strategies, resuming from checkpoints (\texttt{--resume\_from}), and multi-node training. Training outputs are organized into structured directories for checkpoints, samples, results, and logs. This modular execution system enables efficient, reproducible training and evaluation.

\subsection{Training Strategies}
\subsubsection{Layer Freezing}
Layer freezing is a PEFT strategy that selectively disables gradient updates for certain layers of a pre-trained model \cite{layer_freeze}. This is valuable when adapting large VFMs like DINOv2 with limited computational resources.

The principle stems from the observation that not all layers contribute equally; some behave almost as identity mappings and are prime candidates for freezing \cite{layer_freeze}. This offers several advantages:
\begin{itemize}
    \item \textbf{Reduced computational cost:} Prevents backpropagation through frozen layers.
    \item \textbf{Lower memory requirements:} No gradients are stored for frozen layers.
    \item \textbf{Prevention of catastrophic forgetting:} Preserves valuable pre-trained representations.
    \item \textbf{Accelerated convergence:} Fewer parameters to optimize reduces training time.
\end{itemize}

In DINO-MX, the first $N$ layers of the pre-trained model are frozen by default, based on findings that early layers capture general features that transfer well. The value of $N$ is configurable. This provides an effective balance between computational efficiency and downstream performance, and can be combined with other PEFT techniques like LoRA.

\subsubsection{Model Distillation}
The DINO training technique is a form of self-distillation where identical models train each other. The student model approximates the teacher's representations, and the teacher is updated via an Exponential Moving Average (EMA) of the student's weights. This allows the teacher to train stably while the student aims to match the knowledge distilled from itself, producing generalized representations \cite{distill_paper}.

\begin{figure}[htbp]
\centering
\includegraphics[width=1\textwidth]{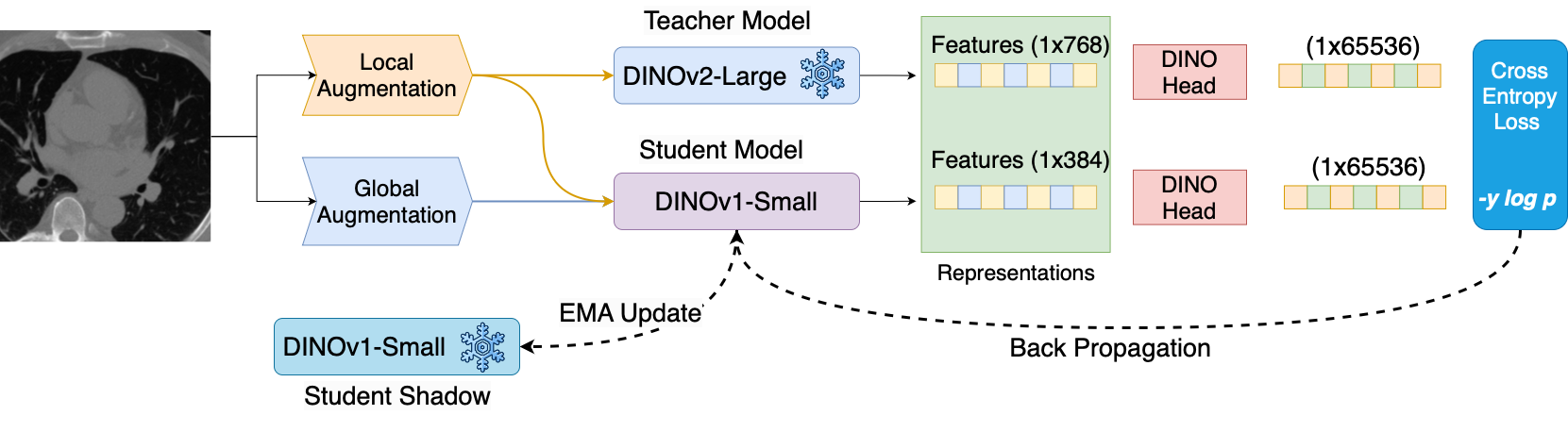}
\caption{Example representation of knowledge distillation system in DINO-MX framework.}
\label{fig:dino_distill}
\end{figure}

Training large foundational models from scratch is resource-intensive. In cases of insufficient GPU power, it is possible to distill knowledge from these large-scale models to smaller ones. The DINO-MX framework employs a knowledge distillation technique based on the DINO methodology \cite{knowledge_dist}. This enables knowledge transfer from large, domain-specific models to smaller 'base' or 'small' models.

Figure \ref{fig:dino_distill} illustrates this process. A larger model (e.g., DINOv2-Large) serves as the teacher, and a smaller model (e.g., DINOv1-Small) acts as the student. The teacher receives locally augmented views, while the student receives both global and local views. The teacher extracts higher-dimensional features (e.g., $1 \times 768$) and the student lower-dimensional ones (e.g., $1 \times 384$). Both process these features through DINO heads to generate equal-dimension outputs (e.g., $1 \times 65536$). Knowledge is transferred via a Cross-Entropy Loss. Importantly, the teacher is not updated via backpropagation, but through an EMA update from a "student shadow" model, ensuring stable knowledge transfer.

\subsubsection{LoRA Adaptation}
Low-Rank Adaptation (LoRA) is an efficient approach for adapting pre-trained VFMs like DINOv2 by significantly reducing the number of trainable parameters \cite{lora_paper}. This PEFT technique is valuable for large models where conventional fine-tuning is computationally prohibitive.

LoRA works by freezing the pre-trained weights ($W_0$) and injecting trainable low-rank decomposition matrices ($B$ and $A$) into specific layers, typically the attention and feed-forward blocks. The weight update $\Delta W$ is parameterized as a low-rank decomposition:

\begin{equation}\label{eq:lora_update}
W = W_0 + \Delta W = W_0 + BA
\end{equation}

where $B \in \mathbb{R}^{d \times r}$ and $A \in \mathbb{R}^{r \times k}$ are the trainable matrices, and the rank $r \ll \min(d,k)$ (typically 4-64).

During the forward pass, the transformation is modified as:
\begin{equation}\label{eq:lora_forward}
h = W_0x + \Delta Wx = W_0x + BAx = W_0x + B(Ax)
\end{equation}

This allows for efficient computation, as matrix $A$ first reduces the dimension of input $x$ to $r$. When applying LoRA to DINOv2, it is typically focused on the self-attention query (Q) and value (V) projection matrices.

This approach reduces trainable parameters by orders of magnitude, requires less memory, eliminates catastrophic forgetting, and enables efficient task switching by only swapping the small adapter weights ($B$ and $A$) while sharing the large frozen backbone.

\subsubsection{Medical Data Augmentation and Label-Guided Training}
\paragraph{Data Augmentation.}
Data augmentation techniques are key components for SSL to enhance robustness and improve generalizable representations \cite{general_aug}. Common techniques include random cropping, flipping, rotating, and color jittering.

However, studies using medical images (e.g., CT scans) often exclude augmentations like solarization and color jitter, as they are unsuitable for this modality \cite{65_aug}. Instead, alternatives like noise addition and brightness changing are used. Table \ref{tab:data_aug} compares common augmentations for RGB and medical images, as outlined in our approach \cite{medical_aug}.

\begin{table}[!ht]
\caption{Comparison of augmentation techniques utilized for RGB images and medical images. (\texttt{amssymb} package required for \cmark and \xmark).}
\label{tab:data_aug}
\centering
\begin{tabular}{lcc} 
\toprule
\textbf{Augmentation Tech.} & \textbf{RGB Images} & \textbf{Medical Images} \\
\midrule 
Random Horizontal Flip & \cmark & \cmark \\
Random Vertical Flip & \cmark & \cmark \\
Random Crop & \cmark & \cmark \\
Random Resized Crop & \cmark & \cmark \\
Gaussian Blur & \cmark & \cmark \\
Solarization & \cmark & \xmark \\ 
Color Jitter & \cmark & \xmark \\ 
Brightness Changing & \xmark & \cmark \\ 
Noise Addition & \xmark & \cmark \\ 
\bottomrule 
\end{tabular}
\end{table} 

\paragraph{Label-Guided Data Augmentation.}
Random augmentation, especially random cropping, significantly improves performance on classification tasks, particularly with limited datasets \cite{survey_data_aug}. However, as dataset size grows, models trained with SSL tend to generate more generalized representations, which can lead to performance decreases in specific tasks \cite{fm_performance}.

The DINO training approach includes a parameter for the number of random local crops (typically 0.05 to 0.4 ratio of the image), which are crucial for generating generalized representations. To make this training more efficient and focus on targeted areas (e.g., calcified ROIs), we introduce \textbf{Label-guided data augmentation}.

\begin{figure}[htbp]
    \centering
    \includegraphics[width=\textwidth]{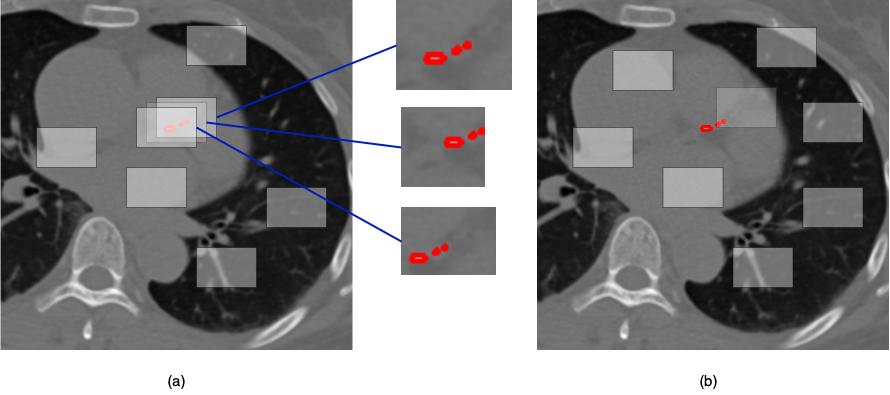}
    \caption{Representation for label guided augmentation, focusing on specific ROIs and generating more than one cropping for calcified regions (a). Completely random local image cropping is utilized by standard DINO training (b).} 
    \label{fig:label_guided_illustration} 
\end{figure}

This technique uses the same number of local crops as the standard approach. In addition to these random crops, it checks for corresponding labels for the input images. These labels are used to *force* the model to learn specific local features by generating additional crops centered on annotated regions. This is depicted in Figure \ref{fig:label_guided_illustration}. These label-guided crops also have the medical augmentation techniques from Table \ref{tab:data_aug} applied to them. By randomly selecting pixel coordinates from labels as "focusing centers," the model is guided to focus on specific regions, preventing it from learning unnecessary local features.

\subsubsection{Attention Map Based Analysis}
ViTs offer valuable interpretability through attention visualization. The self-attention mechanism provides a natural way to visualize which parts of an image the model focuses on. These attention maps offer insights into the model's internal representations.

The attention map is derived from the Multi-Head Self-Attention (MHSA) mechanism. For an input sequence of patch embeddings $\mathbf{X} \in \mathbb{R}^{n \times d}$, the attention weights are computed as:

\begin{equation}
\mathbf{A} = \text{softmax}\left(\frac{\mathbf{Q}\mathbf{K}^T}{\sqrt{d_k}}\right)
\end{equation}

where $\mathbf{Q} = \mathbf{X}\mathbf{W}_Q$, $\mathbf{K} = \mathbf{X}\mathbf{W}_K$ are the query and key projections, and $d_k$ is the dimension of the key vectors.

For example, a $512 \times 512$ image with a $16 \times 16$ patch size results in $32 \times 32 = 1024$ patches. Including the CLS token, $n=1025$, resulting in an attention matrix $\mathbf{A} \in \mathbb{R}^{1025 \times 1025}$. For visualization, we typically extract the attention from the CLS token to all image patches, $\mathbf{M} = \mathbf{A}[0, 1:] \in \mathbb{R}^{1024}$. This reshapes to a $32 \times 32$ grid. With 12 attention heads, we obtain 12 such maps.

\begin{figure}[!ht]
\centering
\includegraphics[width=0.9\linewidth]{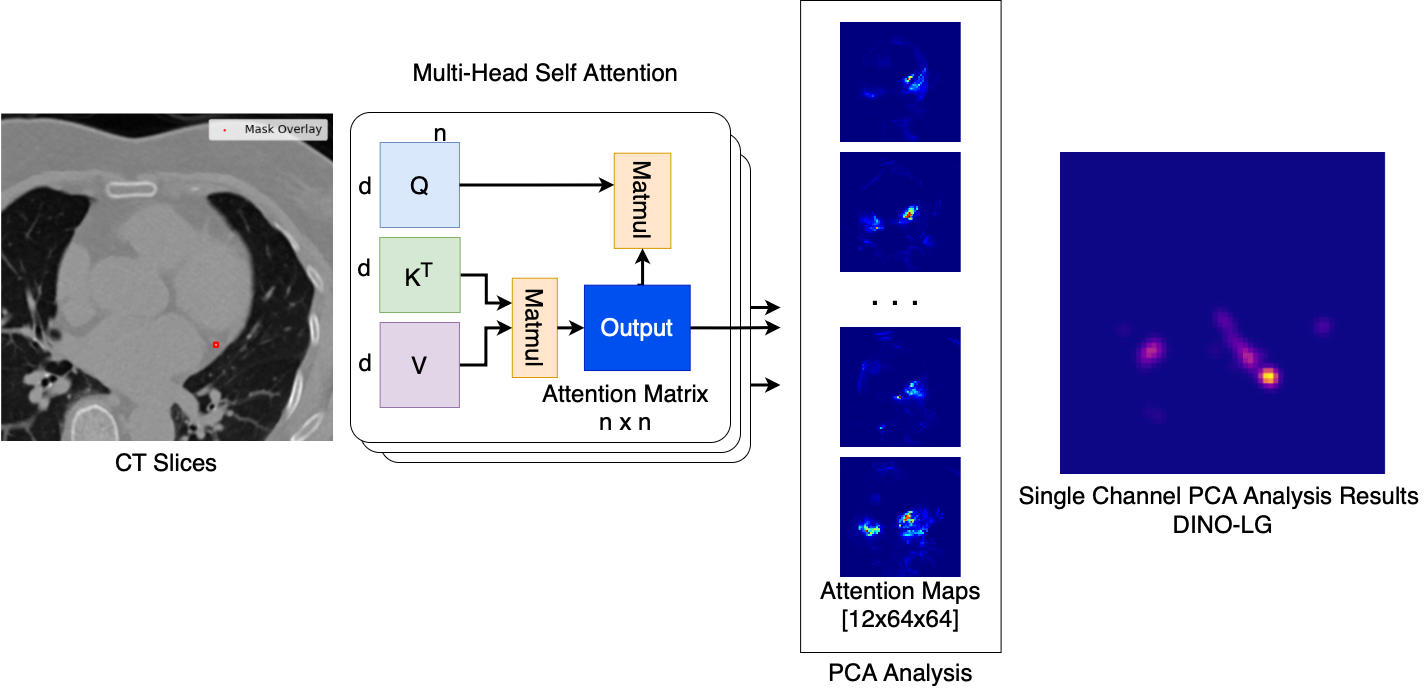}
\caption{Visualization of attention from the CLS token to image patches. For a $512 \times 512$ image with $16 \times 16$ patches, each of the 12 attention heads produces a $32 \times 32$ attention map.}
\label{fig:vit_vis}
\end{figure}

However, direct visualization of MHSA outputs may not be sufficient. Recent studies reveal these maps capture significant features usable for classification and segmentation \cite{att_analysis}. To address this, DINO-MX integrates a PCA analysis tool (Figure \ref{fig:vit_vis}) that conducts Principal Component Analysis on MHSA outputs, reducing the dimensionality from 12 heads to a desired number of channels. This allows researchers to extract and visualize the most significant attention patterns, making it easier to interpret the model's focus.

\subsection{Modularity \& Flexibility}
Modularity in DINO-MX refers to its design of distinct, interchangeable components. This enables easier maintenance, flexibility, reusability, scalability, and parallel development.

In the DINO-MX framework, any ViT model defined in a configuration file can be trained with any technique, including DINOv1 and DINOv2. This flexibility allows researchers to extend their ablation and experimental studies by testing unusual combinations. The models chosen as backbones are standardized (currently limited to Huggingface models), which also provides advantages in data augmentation, as standard inputs and outputs allow for a modular system. This makes it possible to fine-tune models originally trained with natural data augmentations using task-specific ones, like our label-guided medical augmentation.

\subsubsection{Parallelization}
Parallel computation is a necessity for training foundational models. DINO-MX implements both DDP and FSDP. DDP is the conventional approach, maintaining identical model copies on all devices, but creates memory bottlenecks for large models. FSDP addresses this by sharding model parameters, gradients, and optimizer states across devices, allowing much larger models to be trained on the same hardware \cite{FSD_paper, DDP_paper}.

In current frameworks like DINOv1 (DDP) and DINOv2 (FSDP), these techniques are hardcoded. DINO-MX provides a flexible approach, allowing researchers to run all DINO model types with *both* parallelization techniques. This enables comprehensive benchmarking to find optimal configurations for specific medical imaging applications.

\subsubsection{Cross Training}
Recent SSL frameworks often use ViT architectures defined within their own implementation, creating compatibility challenges. Models trained with standard DINOv1/v2 releases may require significant reshaping to be made public or interoperable with systems like Huggingface.

To address this, DINO-MX builds ViT models *directly* on the Huggingface \texttt{transformers} library, ensuring complete compatibility. All ViT specifications are modifiable through configuration files, eliminating code-level changes. This adaptation allows any transformer-based ViT model to be trained with either DINOv1 or DINOv2 techniques. Researchers can now evaluate previously impossible combinations (e.g., fine-tuning a DINOv1 model with DINOv2's iBOT patch loss), bridging the technical gaps between different training methodologies.

\section{Experiments and Results}
\label{sec:experiments}

All experiments were conducted using a standardized hardware configuration consisting of two NVIDIA A6000 GPUs. Models were typically trained for 2000 steps with a batch size of 64 per GPU. For the medical image classification tasks, we utilized the MedMNIST v2 dataset collection \cite{medmnist}, specifically focusing on BloodMNIST (microscopic blood cell images), PathMNIST (colorectal cancer pathology slides), DermaMNIST (dermatoscopic skin lesion images), and OrganAMNIST (axial CT slices of abdominal organs). The attention map detection experiments were conducted on a specialized CT calcification dataset with pixel-level annotations, while model distillation experiments leveraged pathology images compatible with the Prov-Gigapath teacher model. For each configuration, we maintained consistent hyperparameters to ensure fair comparisons, with the primary variable being the adaptation method (standard fine-tuning, LoRA, or layer freezing).

\subsection{Evaluation Metrics}
\begin{figure}[htbp]
\centering
\includegraphics[width=0.8\textwidth]{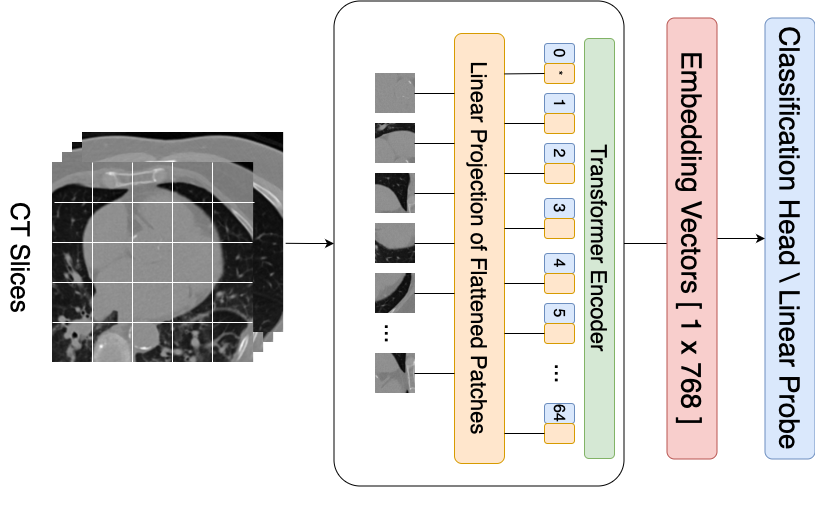}
\caption{General representation of linear probe utilization in DINO-MX framework.}
\label{fig:linear_probe}
\end{figure}

The framework implements a comprehensive evaluation strategy. Linear Probe evaluation serves as a primary metric, training a single linear layer on frozen representations, as illustrated in Figure~\ref{fig:linear_probe}. This quantifies the linear separability of the learned features without modifying the representation space \cite{vfm_class}.

\begin{figure}[htbp]
\centering
\includegraphics[width=1\textwidth]{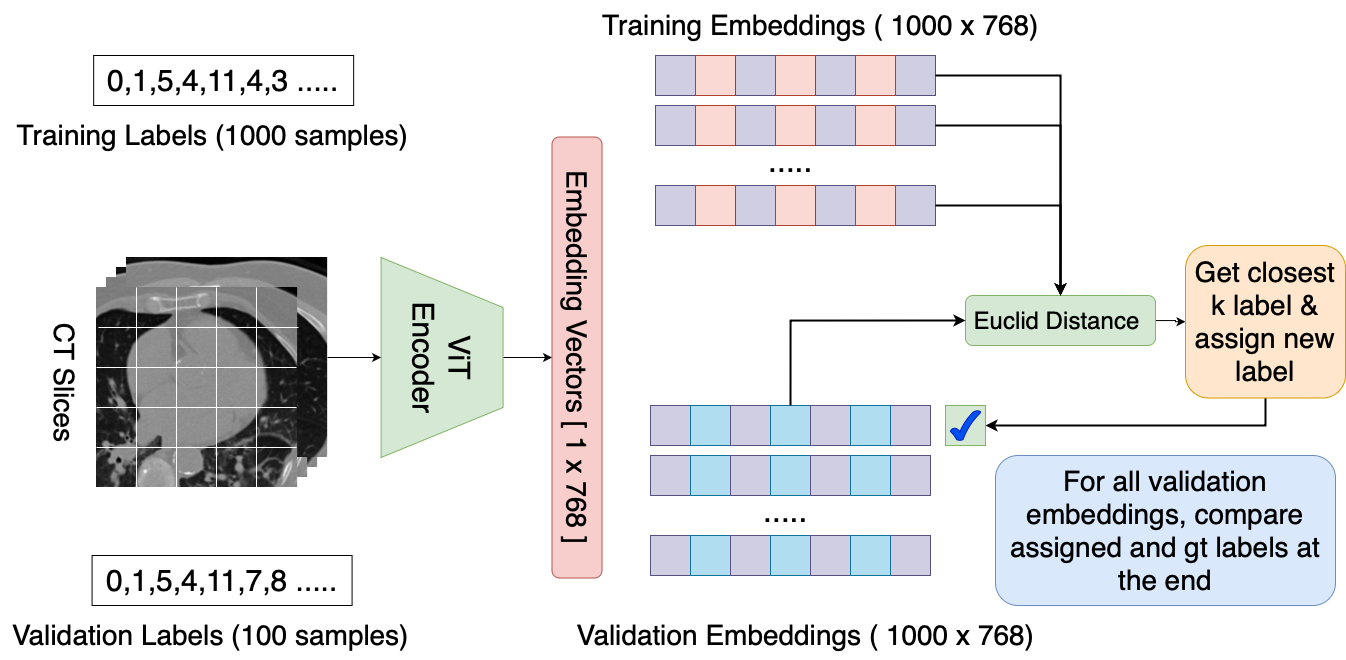}
\caption{Representation of KNN training for evaluating the embedding vectors.}
\label{fig:knn_training}
\end{figure}

Complementing this, k-Nearest Neighbors (k-NN) \cite{knn_paper} classification offers a non-parametric evaluation that directly measures the geometric quality of the embedding space, detailed in Figure~\ref{fig:knn_training}. For each test sample, we compute pairwise distances to all training embeddings and identify the $k$ nearest samples (typically $k=5$) for a majority vote.

Across all experiments, k-NN classification consistently outperforms linear probing by substantial margins (10-25\% on BloodMNIST and PathMNIST), demonstrating that DINO-MX learns representations with complex non-linear relationships well-suited for neighborhood-based decision boundaries.

\subsection{Experiments on Attention Map Based Detection and DINO-LG}
Experiments with DINO-LG (Label-Guided) demonstrate how this technique leverages annotated labels as guidance during data augmentation. This trains the model to focus on specific regions of interest (e.g., calcifications in CT) without needing separate detection heads. By applying PCA and clustering to the resulting attention maps, we can effectively identify highlighted regions.

\begin{figure}[!ht]
\centering
\includegraphics[width=1\linewidth]{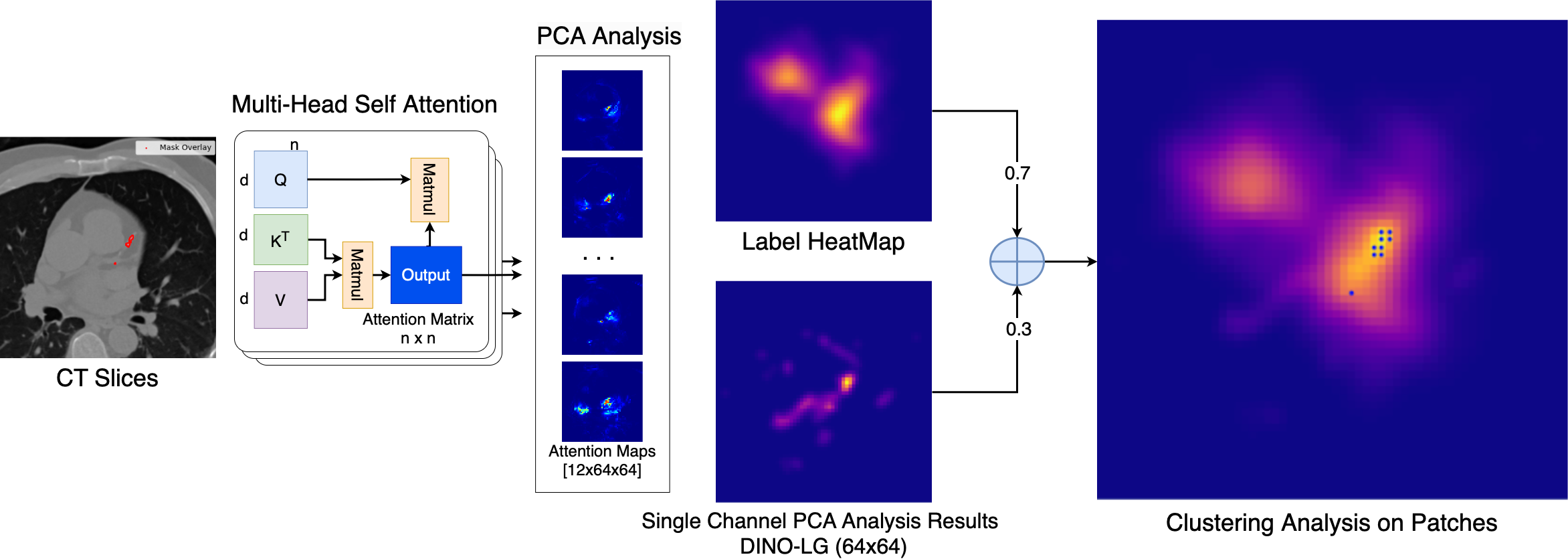}
\caption{Attention-based detection pipeline for calcifications in CT slices using DINO-LG. PCA is applied to attention maps, and clustering analysis on patches with attention values above thresholds (0.3 and 0.7) identifies regions of interest.}
\label{fig:att_set}
\end{figure}

Figure~\ref{fig:att_set} illustrates the full pipeline. The MHSA mechanism generates $12 \times 64 \times 64$ attention maps. PCA extracts significant features, producing single-channel results that correlate with ground truth heatmaps. By thresholding (0.3 and 0.7) and clustering, we identify calcified patches.

The results in Table~\ref{tab:dino_lg_results} demonstrate the effectiveness of DINO-LG for calcification detection. It achieves a high average precision (0.7961) and recall (0.7072), with a balanced F1-score (0.7141). The high localization score (0.9010) is particularly noteworthy, indicating accurate localization. Statistics show 1674 true positives against 685 false negatives and 628 false positives. This validates that our label-guided attention mechanism effectively focuses on relevant regions while maintaining computational efficiency.

\begin{table}[!ht]
\caption{Detection Performance of DINO-LG on CT Calcification Dataset}
\label{tab:dino_lg_results}
\centering
\begin{tabular}{lc}
\toprule
\textbf{Metric} & \textbf{Average} \\
\midrule
Precision & 0.7961 \\
Recall & 0.7072 \\
F1-score & 0.7141 \\ 
Localization Score & 0.9010 \\
\midrule
\multicolumn{2}{c}{\textbf{Total Detection Statistics}} \\
\midrule
True Positives & 1674.0 \\
False Positives & 628.0 \\
False Negatives & 685.0 \\
\bottomrule
\end{tabular}
\end{table}

\subsection{Experiments on Medical Image Classification}
We conducted extensive experiments on the MedMNIST v2 dataset \cite{medmnist}, focusing on BloodMNIST, PathMNIST, DermaMNIST, and OrganAMNIST. This diverse set of microscopic and macroscopic images allows a comprehensive evaluation of generalizability.

Our experimental design assessed:
\begin{itemize}
    \item Performance comparison between DINOv1 and DINOv2 models.
    \item Impact of PEFT strategies (LoRA, layer freezing) on accuracy and efficiency.
    \item Evaluation of classification approaches (Linear Probe vs. k-NN).
    \item Cross-training capabilities (e.g., DINOv1 model with DINOv2 technique).
    \item Adaptation performance across all four tasks.
\end{itemize}

We evaluated DINOv1 (dino-vitb8, dino-vitb16) and DINOv2 (dinov2-small, dinov2-base, dinov2-large) models with standard fine-tuning, LoRA, or layer freezing.

\begin{table}[!ht]
\caption{BloodMNIST Classification Performance}
\label{tab:bloodmnist_results}
\centering
\resizebox{\textwidth}{!}{
\begin{tabular}{lccccccccc}
\toprule
& \multicolumn{4}{c}{\textbf{Linear Probe}} & \multicolumn{4}{c}{\textbf{k-Nearest Neighbors}} \\
\cmidrule(lr){2-5} \cmidrule(lr){6-9}
\textbf{Model} & \textbf{Adaptation} & \textbf{Accuracy} & \textbf{Precision} & \textbf{F1} & \textbf{Adaptation} & \textbf{Accuracy} & \textbf{Precision} & \textbf{F1} \\
\midrule
\multirow{4}{*}{dino-vits16} 
& Pretrained & 0.53 & 0.44 & 0.42 & Pretrained & 0.97 & 0.97 & 0.97 \\
& Standard & \textbf{0.91} & 0.91 & 0.90 & Standard & 0.99 & 0.99 & 0.99 \\
& LoRA & 0.67 & 0.80 & 0.60 & LoRA & 0.98 & 0.98 & 0.98 \\
& Layer Freezing & 0.89 & 0.89 & 0.88 & Layer Freezing & 0.99 & 0.99 & 0.99 \\
\midrule
\multirow{4}{*}{dino-vitb16} 
& Pretrained & 0.71 & 0.77 & 0.66 & Pretrained & 0.97 & 0.97 & 0.97 \\
& Standard & \textbf{0.95} & 0.95 & 0.95 & Standard & 0.99 & 0.99 & 0.99 \\
& LoRA & 0.81 & 0.85 & 0.79 & LoRA & 0.99 & 0.99 & 0.99 \\
& Layer Freezing & \textbf{0.95} & 0.95 & 0.95 & Layer Freezing & 0.99 & 0.99 & 0.99 \\
\midrule
\multirow{4}{*}{dinov2-small} 
& Pretrained & 0.34 & 0.30 & 0.24 & Pretrained & 0.94 & 0.94 & 0.94 \\
& Standard & 0.93 & 0.94 & 0.93 & Standard & 0.99 & 0.99 & 0.99 \\
& LoRA & 0.54 & 0.56 & 0.47 & LoRA & 0.98 & 0.98 & 0.98 \\
& Layer Freezing & \textbf{0.94} & 0.94 & 0.94 & Layer Freezing & 0.99 & 0.99 & 0.99 \\
\midrule
\multirow{4}{*}{dinov2-base} 
& Pretrained & 0.54 & 0.63 & 0.47 & Pretrained & 0.94 & 0.94 & 0.94 \\
& Standard & \textbf{0.95} & 0.95 & 0.95 & Standard & 0.99 & 0.99 & 0.99 \\
& LoRA & 0.80 & 0.82 & 0.75 & LoRA & 0.98 & 0.98 & 0.98 \\
& Layer Freezing & \textbf{0.95} & 0.95 & 0.95 & Layer Freezing & 0.99 & 0.99 & 0.99 \\
\bottomrule
\end{tabular}
}
\end{table}

\begin{table}[!ht]
\caption{DermaMNIST Classification Performance}
\label{tab:dermamnist_results}
\centering
\resizebox{\textwidth}{!}{
\begin{tabular}{lccccccccc}
\toprule
& \multicolumn{4}{c}{\textbf{Linear Probe}} & \multicolumn{4}{c}{\textbf{k-Nearest Neighbors}} \\
\cmidrule(lr){2-5} \cmidrule(lr){6-9}
\textbf{Model} & \textbf{Adaptation} & \textbf{Accuracy} & \textbf{Precision} & \textbf{F1} & \textbf{Adaptation} & \textbf{Accuracy} & \textbf{Precision} & \textbf{F1} \\
\midrule
\multirow{4}{*}{dino-vits16} 
& Pretrained & 0.66 & 0.43 & 0.52 & Pretrained & 0.79 & 0.77 & 0.77 \\
& Standard & \textbf{0.67} & 0.62 & 0.64 & Standard & 0.78 & 0.77 & 0.77 \\
& LoRA & 0.66 & 0.43 & 0.52 & LoRA & 0.78 & 0.77 & 0.77 \\
& Layer Freezing & \textbf{0.67} & 0.60 & 0.63 & Layer Freezing & 0.77 & 0.76 & 0.76 \\
\midrule
\multirow{4}{*}{dino-vitb16} 
& Pretrained & 0.67 & 0.49 & 0.55 & Pretrained & 0.79 & 0.77 & 0.77 \\
& Standard & \textbf{0.71} & 0.62 & 0.65 & Standard & 0.80 & 0.79 & 0.79 \\
& LoRA & 0.66 & 0.54 & 0.52 & LoRA & 0.79 & 0.77 & 0.77 \\
& Layer Freezing & 0.69 & 0.58 & 0.62 & Layer Freezing & 0.79 & 0.78 & 0.78 \\
\midrule
\multirow{4}{*}{dinov2-small} 
& Pretrained & 0.66 & 0.48 & 0.53 & Pretrained & 0.76 & 0.73 & 0.74 \\
& Standard & \textbf{0.69} & 0.61 & 0.62 & Standard & 0.76 & 0.73 & 0.74 \\
& LoRA & 0.65 & 0.50 & 0.55 & LoRA & 0.76 & 0.75 & 0.74 \\
& Layer Freezing & \textbf{0.69} & 0.63 & 0.62 & Layer Freezing & 0.76 & 0.74 & 0.74 \\
\midrule
\multirow{4}{*}{dinov2-base} 
& Pretrained & 0.68 & 0.59 & 0.62 & Pretrained & 0.76 & 0.74 & 0.73 \\
& Standard & \textbf{0.71} & 0.66 & 0.68 & Standard & 0.75 & 0.74 & 0.74 \\
& LoRA & 0.69 & 0.66 & 0.65 & LoRA & 0.76 & 0.73 & 0.73 \\
& Layer Freezing & \textbf{0.71} & 0.65 & 0.67 & Layer Freezing & 0.76 & 0.74 & 0.74 \\
\bottomrule
\end{tabular}
}
\end{table}

\begin{table}[!ht]
\caption{OrganAMNIST Classification Performance}
\label{tab:organamnist_results}
\centering
\resizebox{\textwidth}{!}{
\begin{tabular}{lccccccccc}
\toprule
& \multicolumn
{4}{c}{\textbf{Linear Probe}} & \multicolumn{4}{c}{\textbf{k-Nearest Neighbors}} \\
\cmidrule(lr){2-5} \cmidrule(lr){6-9}
\textbf{Model} & \textbf{Adaptation} & \textbf{Accuracy} & \textbf{Precision} & \textbf{F1} & \textbf{Adaptation} & \textbf{Accuracy} & \textbf{Precision} & \textbf{F1} \\
\midrule
\multirow{4}{*}{dino-vits16} 
& Pretrained & 0.57 & 0.59 & 0.53 & Pretrained & 0.95 & 0.95 & 0.95 \\
& Standard & \textbf{0.85} & 0.85 & 0.81 & Standard & 0.97 & 0.97 & 0.97 \\
& LoRA & 0.61 & 0.67 & 0.57 & LoRA & 0.96 & 0.96 & 0.96 \\
& Layer Freezing & \textbf{0.85} & 0.85 & 0.81 & Layer Freezing & 0.96 & 0.97 & 0.96 \\
\midrule
\multirow{4}{*}{dino-vitb16} 
& Pretrained & 0.75 & 0.80 & 0.72 & Pretrained & 0.96 & 0.96 & 0.96 \\
& Standard & \textbf{0.92} & 0.93 & 0.92 & Standard & 0.97 & 0.98 & 0.97 \\
& LoRA & 0.78 & 0.82 & 0.76 & LoRA & 0.97 & 0.97 & 0.97 \\
& Layer Freezing & 0.91 & 0.91 & 0.90 & Layer Freezing & 0.97 & 0.97 & 0.97 \\
\midrule
\multirow{4}{*}{dinov2-small} 
& Pretrained & 0.64 & 0.65 & 0.61 & Pretrained & 0.94 & 0.95 & 0.94 \\
& Standard & \textbf{0.86} & 0.88 & 0.84 & Standard & 0.96 & 0.97 & 0.96 \\
& LoRA & 0.71 & 0.70 & 0.68 & LoRA & 0.96 & 0.96 & 0.96 \\
& Layer Freezing & 0.83 & 0.84 & 0.82 & Layer Freezing & 0.96 & 0.96 & 0.96 \\
\midrule
\multirow{4}{*}{dinov2-base} 
& Pretrained & 0.75 & 0.77 & 0.74 & Pretrained & 0.92 & 0.92 & 0.92 \\
& Standard & \textbf{0.88} & 0.89 & 0.87 & Standard & 0.97 & 0.97 & 0.97 \\
& LoRA & 0.80 & 0.83 & 0.79 & LoRA & 0.93 & 0.94 & 0.93 \\
& Layer Freezing & \textbf{0.88} & 0.89 & 0.87 & Layer Freezing & 0.97 & 0.97 & 0.97 \\
\bottomrule
\end{tabular}
}
\end{table}

\begin{table}[!ht]
\caption{PathMNIST Classification Performance}
\label{tab:pathmnist_results}
\centering
\resizebox{\textwidth}{!}{
\begin{tabular}{lccccccccc}
\toprule
& \multicolumn{4}{c}{\textbf{Linear Probe}} & \multicolumn{4}{c}{\textbf{k-Nearest Neighbors}} \\
\cmidrule(lr){2-5} \cmidrule(lr){6-9}
\textbf{Model} & \textbf{Adaptation} & \textbf{Accuracy} & \textbf{Precision} & \textbf{F1} & \textbf{Adaptation} & \textbf{Accuracy} & \textbf{Precision} & \textbf{F1} \\
\midrule
\multirow{4}{*}{dino-vits16} 
& Pretrained & 0.73 & 0.75 & 0.72 & Pretrained & 0.99 & 0.99 & 0.99 \\
& Standard & \textbf{0.91} & 0.91 & 0.91 & Standard & 0.99 & 0.99 & 0.99 \\
& LoRA & 0.76 & 0.79 & 0.74 & LoRA & 0.99 & 0.99 & 0.99 \\
& Layer Freezing & 0.89 & 0.89 & 0.89 & Layer Freezing & 0.99 & 0.99 & 0.99 \\
\midrule
\multirow{4}{*}{dino-vitb16} 
& Pretrained & 0.88 & 0.88 & 0.88 & Pretrained & 0.99 & 0.99 & 0.99 \\
& Standard & \textbf{0.94} & 0.94 & 0.94 & Standard & 0.99 & 0.99 & 0.99 \\
& LoRA & 0.89 & 0.90 & 0.89 & LoRA & \textbf{1.00} & 1.00 & 1.00 \\
& Layer Freezing & \textbf{0.94} & 0.95 & 0.94 & Layer Freezing & \textbf{1.00} & 1.00 & 1.00 \\
\midrule
\multirow{4}{*}{dinov2-small} 
& Pretrained & 0.62 & 0.63 & 0.62 & Pretrained & 0.98 & 0.98 & 0.98 \\
& Standard & \textbf{0.92} & 0.92 & 0.92 & Standard & 0.99 & 0.99 & 0.99 \\
& LoRA & 0.83 & 0.84 & 0.83 & LoRA & 0.99 & 0.99 & 0.99 \\
& Layer Freezing & \textbf{0.92} & 0.92 & 0.92 & Layer Freezing & 0.99 & 0.99 & 0.99 \\
\midrule
\multirow{4}{*}{dinov2-base} 
& Pretrained & 0.76 & 0.78 & 0.75 & Pretrained & 0.98 & 0.98 & 0.98 \\
& Standard & \textbf{0.95} & 0.95 & 0.95 & Standard & 0.99 & 0.99 & 0.99 \\
& LoRA & 0.83 & 0.86 & 0.80 & LoRA & 0.99 & 0.99 & 0.99 \\
& Layer Freezing & \textbf{0.95} & 0.95 & 0.95 & Layer Freezing & 0.99 & 0.99 & 0.99 \\
\bottomrule
\end{tabular}
}
\end{table}

Tables \ref{tab:bloodmnist_results}, \ref{tab:dermamnist_results}, \ref{tab:organamnist_results}, and \ref{tab:pathmnist_results} present the detailed classification performance. Several consistent patterns emerge:
\begin{itemize}
    \item DINOv2 and DINOv1 models perform comparably, with no architecture showing consistent dominance.
    \item Highest performance is on PathMNIST (Table \ref{tab:pathmnist_results}), where dino-vitb16 achieves 100\% kNN accuracy with LoRA and Layer Freezing, likely due to distinctive histological patterns.
    \item BloodMNIST (Table \ref{tab:bloodmnist_results}) also shows strong kNN performance, suggesting morphological features are well-captured.
    \item DermaMNIST (Table \ref{tab:dermamnist_results}) shows the lowest performance (max 80\% kNN accuracy), reflecting the challenge of subtle skin texture differences.
    \item kNN classification substantially outperforms linear probing across all datasets, especially on BloodMNIST and PathMNIST (e.g., 70-80\% linear probe to 97-100\% kNN). This suggests the learned features contain rich, non-linear structures that linear classifiers cannot capture.
\end{itemize}

\begin{table}[!ht]
\caption{Comparison of DINO-MX with State-of-the-Art Models on MedMNIST Datasets.}
\label{tab:benchmark_comparison}
\centering
\begin{tabular}{lcccc}
\toprule
\textbf{Dataset} & \multicolumn{2}{c}{\textbf{Benchmark Models}} & \multicolumn{2}{c}{\textbf{DINO-MX (Linear Probe)}} \\
\cmidrule(lr){2-3} \cmidrule(lr){4-5}
& \textbf{Best Model} & \textbf{Accuracy} & \textbf{Best Model} & \textbf{Accuracy} \\
\midrule
BloodMNIST & Google AutoML Vision & 0.966 & dino-vitb16 (Standard) & 0.950 \\
PathMNIST & ResNet-50 (28) & 0.911 & dinov2-base (Standard) & 0.950 \\
DermaMNIST & Google AutoML Vision & 0.768 & dinov2-base (Standard) & 0.710 \\
OrganAMNIST & ResNet-18 (224) & 0.951 & dino-vitb16 (Standard) & 0.920 \\
\bottomrule
\multicolumn{5}{p{0.9\textwidth}}{\small Note: All DINO-MX models were trained for 2000 steps using 2 NVIDIA A6000 GPUs with a batch size of 64 per GPU.} \\
\end{tabular}
\end{table}

As shown in Table \ref{tab:benchmark_comparison}, DINO-MX demonstrates competitive performance with SOTA benchmarks even using only linear probing. Our dinov2-base model achieves 95.0\% on PathMNIST, significantly outperforming ResNet-50 (91.1\%). On BloodMNIST and DermaMNIST, our models approach SOTA results. This is particularly impressive given the computational efficiency of our PEFT strategies.

\begin{table}[!ht]
\caption{Computational Efficiency of Different Adaptation Strategies.}
\label{tab:computational_efficiency}
\centering
\begin{tabular}{lccc}
\toprule
\textbf{Model} & \textbf{Adaptation Strategy} & \textbf{Training Time (mins)} & \textbf{Memory Usage ( GB $\times$ \# GPU )} \\
\midrule
\multirow{3}{*}{dino-vitb16} & Standard & 25 & 19.3 \\
& LoRA & 19 & 15.0 \\
& Layer Freezing (6 Layers) & 18 & 16.2 \\
\midrule
\multirow{3}{*}{dinov2-base} & Standard & 37 & 28.5 \\
& LoRA & 28 & 22.3 \\
& Layer Freezing (6 Layers) & 23.8 & 25 \\
\bottomrule
\end{tabular}
\end{table}

When comparing adaptation strategies, standard fine-tuning yields the highest linear probe accuracy, but differences are negligible for kNN. LoRA and layer freezing offer compelling, resource-efficient alternatives. Table \ref{tab:computational_efficiency} demonstrates these advantages. LoRA provides the most significant savings, reducing training time by 35-40\% and memory by 35-36\%, making it ideal for resource-constrained environments. Layer freezing offers a middle ground.

\begin{table}[!ht]
\caption{Cross-Training Results Between DINOv1 and DINOv2 (Linear Probe Accuracy on Pathmnist dataset).}
\label{tab:cross_training}
\centering
\begin{tabular}{lcc}
\toprule
\textbf{Model Configuration} & \textbf{Standard Approach} & \textbf{Cross-Training} \\
\midrule
DINOv1 (vitb16) & 0.94 & 0.93 \\ 
DINOv2 (base) & 0.95 & 0.82 \\ 
\bottomrule
\end{tabular}
\end{table}

A unique capability of DINO-MX is cross-training. Table \ref{tab:cross_training} shows these results. DINOv1 models are resilient when trained with DINOv2 techniques (0.94 $\rightarrow$ 0.93). However, DINOv2 models show a substantial performance drop when trained with DINOv1 methodologies (0.95 $\rightarrow$ 0.82). This asymmetric behavior suggests newer architectures are more optimized for their corresponding training techniques.

In conclusion, our experiments demonstrate DINO-MX's effectiveness and computational advantages. The framework achieves competitive performance, and the significant gap between linear probe and kNN highlights the rich, non-linear representations learned, which are valuable for medical imagery.

\subsection{Experiments on Model Distillation}
Model distillation offers an efficient way to transfer knowledge from large models to smaller ones. We used Prov-Gigapath \cite{prov-gigapath}, a large-scale model trained on pathology images, as a teacher to distill knowledge into two student models: dinov2-small and dino-vits16. We evaluated this on the PathMNIST dataset using three adaptation modes: frozen, LoRA, and standard.

\begin{table}[!ht]
\caption{Comparison of Pretraining and Distillation Performance on PathMNIST.}
\label{tab:pretrain_vs_distill}
\centering
\begin{tabular}{@{\hskip 6pt}ll@{\hskip 6pt}l@{\hskip 6pt}cc@{\hskip 12pt}cc@{\hskip 6pt}}
\toprule
\textbf{Teacher} & \textbf{Student} & \textbf{Strategy} 
& \multicolumn{2}{c}{\textbf{Fine-Tuning}} 
& \multicolumn{2}{c}{\textbf{Distill}} \\
& & & \textbf{Linear Acc.} & \textbf{kNN Acc.} & \textbf{Linear Acc.} & \textbf{kNN Acc.} \\
\midrule
\multicolumn{7}{l}{\textit{Teacher-Only Performance}} \\
Prov-Gigapath \cite{prov-gigapath} & - & - & - & - & 0.97 & 0.99 \\
Uni \cite{uni_paper} & - & - & - & - & 0.97 & 0.99 \\
\midrule
\multicolumn{7}{l}{\textit{Distillation Results}} \\
\multirow{6}{*}{Prov-Gigapath} 
  & \multirow{3}{*}{dinov2-small} 
    & Standard & 0.92 & 0.99 & \textbf{0.96} & 0.99 \\
  & & LoRA & 0.83 & 0.99 & \textbf{0.90} & 0.98 \\
  & & Freeze (6) & 0.92 & 0.99 & \textbf{0.96} & 0.99 \\
\cmidrule{2-7}
  & \multirow{3}{*}{dino-vits16} 
    & Standard & 0.91 & 0.99 & \textbf{0.91} & 0.98 \\
  & & LoRA & 0.76 & 0.99 & \textbf{0.82} & 0.99 \\
  & & Freeze (6) & 0.89 & 0.99 & \textbf{0.91} & 0.98 \\
\midrule
\multirow{6}{*}{Uni} 
  & \multirow{3}{*}{dinov2-small} 
    & Standard & 0.92 & 0.99 & \textbf{0.96} & 0.99 \\
  & & LoRA & 0.83 & 0.99 & \textbf{0.86} & 0.98 \\
  & & Freeze (6) & 0.92 & 0.99 & \textbf{0.96} & 0.99 \\
\cmidrule{2-7}
  & \multirow{3}{*}{dino-vits16} 
    & Standard & 0.91 & 0.99 & \textbf{0.91} & 0.99 \\
  & & LoRA & 0.76 & 0.99 & \textbf{0.80} & 0.99 \\
  & & Freeze (6) & 0.89 & 0.99 & \textbf{0.91} & 0.99 \\
\bottomrule
\end{tabular}
\end{table}

As shown in Table \ref{tab:pretrain_vs_distill}, knowledge distillation consistently outperforms traditional fine-tuning, especially when evaluated with linear probes. Both teacher models (Prov-Gigapath \cite{prov-gigapath} and Uni \cite{uni_paper}) achieved comparable effectiveness.

The performance gap is most pronounced with the dinov2-small architecture, showing a 4 percentage point improvement in linear probe accuracy when distilled. This shows smaller models particularly benefit from distillation. Interestingly, LoRA adaptation, while having lower absolute performance, showed the largest *relative* improvement when comparing distillation to fine-tuning (a 7-point jump for dinov2-small), indicating PEFT methods can effectively leverage distilled knowledge.

For both student architectures, the Layer Freezing strategy (with 6 frozen layers) achieves nearly identical performance to the Standard approach when using distillation, despite using far fewer trainable parameters. This demonstrates that distillation can compensate for the reduced flexibility of PEFT methods, making them even more viable for resource-constrained clinical deployments.


\section{Discussion}
\label{sec:discussion}

The main focus of this research is to address important challenges related to the training and use of vision foundation models (VFMs) through various novel and flexible approaches. The DINO-MX framework introduced here provides a modular and adaptable research environment, enabling researchers to explore different self-supervised learning methods without the usual complexity of managing multiple independent workflows. By simplifying this process, the framework allows for easier comparison, experimentation, and integration of various training methods, ultimately accelerating advancements in medical imaging.

\paragraph{Domain Adaptation}
One significant issue we aim to solve involves effectively adapting VFMs initially trained on natural images to specific medical imaging contexts. Medical images, such as CT scans and histopathology slides, differ greatly from natural images in terms of pixel intensities, spatial characteristics, and semantic meanings. Developing methods to bridge these domain differences is essential to ensure reliable and accurate results in medical applications. These methods include medical data augmentation techniques and appropriate hyperparameters in training of medical models. It is explicity clear that it is possible to use these method on pre-training a model from strach as it is compatible with single channel data types such as CT slices. Especially, when it is considered even giant models trained on pathology images utilize standard data augmentation applied on pre-training process, our approach may contribute a significant difference in this field.

\paragraph{Resource-Efficient Training}
Given the high computational costs associated with vision foundational models, another core aspect of our research is exploring efficient training methods. Techniques like Layer Freezing, Low Rank Adaptation (LoRA), and model distillation are investigated to reduce resource consumption significantly. Such approaches make advanced vision models more accessible, particularly benefiting institutions with limited computational resources.

\paragraph{Medical-Specific Data Augmentation}
Traditional data augmentation strategies, typically designed for natural images, often fall short when applied directly to medical datasets. Therefore, a major focus of this research involves the creation and evaluation of data augmentation methods specifically tailored for medical images. These medical-specific techniques are designed to improve the robustness of the model and improve performance in clinical scenarios.

\paragraph{Modular \& Flexible Framework}
The research also introduces the DINO-MX framework, intentionally designed with modularity and flexibility. The framework standardizes the use of common backbone architectures, such as Vision Transformers (ViT), and easily accommodates different medical image types, especially single-channel images like CT scans. Importantly, these ViT models are fully compatible with Hugging Face libraries, ensuring that even modified configurations remain Hugging Face compliant. This compatibility allows researchers to easily share their adapted models publicly,collaborative development and dissemination within the research community. This modular approach ensures data integrity and simplifies the training process without requiring artificial alterations to the input data.

\paragraph{Interpretability and Analysis}
Finally, the interpretability of vision foundational models is crucial for their acceptance in clinical practice. This research emphasizes methods using attention maps to improve transparency and interpretability of model decisions. Such approaches help clinicians better understand and trust the outcomes produced by these advanced models, facilitating their integration into clinical workflows.

By collectively addressing these key challenges, the research aims to significantly improve the performance, efficiency, and practical usability of vision foundational models, promoting broader adoption and meaningful progress in the field of medical imaging.
 
\section{Conclusion and Future Work}
\label{sec:conclusion}

In this research, we identified and analyzed several critical challenges within self-supervised learning frameworks for medical imaging. While recent advancements in foundational models like CT-CLIP and ProvGigapath demonstrate impressive capabilities, they predominantly rely on standard DINOv2 methodologies designed originally for natural images. This discrepancy creates a significant gap between the specialized needs of medical imaging and the general-purpose techniques commonly employed.

The fundamental architectural dependency on frameworks developed for natural images introduces potential mismatches between methodologies and the specific requirements of medical applications. Furthermore, the computational demands of training these models remain prohibitively high, often exceeding the resources available at many research institutions. The limited transparency in the field, exemplified by institutions rarely sharing modifications to training methodologies, further restricts collaborative improvement efforts, as noted in recent repository issues \cite{provgigapath2025issue47}.

To address these challenges, we introduced DINO-MX, a modular and flexible training framework that integrates both DINOv1 and DINOv2 self-supervised learning methods. Specifically tailored for medical imaging while built on established DINO training infrastructure, our framework ensures stability and compatibility with existing research ecosystems. DINO-MX supports extensive experimentation with diverse backbone architectures, offers parameter-efficient fine-tuning through techniques such as LoRA adaptation, provides native support for medical-specific data augmentation strategies, and enables cross-training capabilities among different self-supervised learning paradigms.

It is important to emphasize that our primary objective in developing DINO-MX was not to surpass state-of-the-art performance metrics for any particular model or dataset. Rather, we aimed to create a comprehensive infrastructure that enables researchers, particularly those in medical laboratories, to efficiently experiment across multiple architectures and training methodologies without unnecessary time expenditure. In today's AI landscape, where high-capacity models already demonstrate exceptional capabilities, the critical challenge lies not in achieving marginal performance improvements but in democratizing access to these techniques and accelerating their practical implementation.

Our experimental results across multiple medical imaging datasets confirm this approach, demonstrating that DINO-MX achieves competitive performance compared to state-of-the-art models while significantly reducing computational overhead. This efficiency is particularly valuable in resource-constrained medical environments, where implementation speed and accessibility often outweigh the pursuit of fractional accuracy improvements. The framework's flexibility enables researchers to make informed decisions about model selection based on their specific computational constraints and clinical requirements.

The knowledge distillation capabilities we implemented further enhance this accessibility, allowing smaller, more deployable models to benefit from the representations learned by larger foundational models. This approach strikes an essential balance between computational efficiency and diagnostic performance - a crucial consideration for practical clinical applications where hardware limitations often present significant barriers to adoption.

While DINO-MX primarily focuses on vision models, its impact extends beyond single-modality applications. The vision encoders developed through our framework can serve as critical components within broader multi-modal systems that integrate visual and linguistic understanding. These representations can significantly enhance medical image interpretation when combined with language models, improving diagnostic assistance across various specialties.

Looking forward, we aim to extend DINO-MX by integrating vision encoders with Large Language Models available through the LLM Factory \cite{LLM_factory}. This combination will enable systems that not only see medical images effectively but also reason about them in clinically meaningful ways. By uniting specialized visual representations with the reasoning capabilities of advanced language models, we envision creating interpretable AI systems that can understand, describe, and analyze medical images with contextual awareness that approaches that of human specialists, ultimately providing valuable clinical decision support across diverse medical imaging modalities.
 
\bibliographystyle{unsrt}  
\bibliography{references}

\end{document}